\documentclass[letterpaper, 10 pt, conference]{ieeeconf}  

\IEEEoverridecommandlockouts                              

\usepackage{amsmath} 
\usepackage{color}
\usepackage{soul}
\usepackage{booktabs}
\usepackage{array,makecell}

\usepackage[pdftex]{graphicx}
\graphicspath{./Figs}

\title{\LARGE \bf
Cognitive architecture aided by working-memory for self-supervised multi-modal humans recognition. 
}

\author{Jonas  Gonzalez-Billandon*$^{1}$, Giulia Belgiovine*$^{2}$, Alessandra Sciutti$^{1}$, Giulio Sandini$^{2}$, Francesco Rea$^{2}$ 
\thanks{*Co-authors that contributed equally in the realisation of this paper    {\tt\small jonas.gonzalez@iit.it, giulia.belgiovine@iit.it}}
\thanks{$^{1}$ Cognitive Architecture for Collaborative Technologies lab, Italian Institute of technology, Genova, Italy}%
\thanks{$^{2}$ Robotics Brain and Cognitive Science lab, Italian Institute of technology, Genova, Italy}%
}

\IEEEaftertitletext{ \copyright 2021 IEEE. Personal use of this material is permitted. Permission from IEEE must be
obtained for all other uses, in any current or future media, including reprinting/republishing this material for advertising or promotional purposes, creating new collective works, for resale or redistribution to servers or lists, or reuse of any copyrighted component of this work in other works.}

\begin{document}

\maketitle
\thispagestyle{empty}
\pagestyle{empty}

\begin{abstract}
The ability to recognize human partners is an important social skill to build personalized and long-term human-robot interactions, especially in scenarios like education, care-giving, and rehabilitation. Faces and voices constitute two important sources of information to enable artificial systems to reliably recognize individuals.
Deep learning networks have achieved state-of-the-art results and demonstrated to be suitable tools to address such a task. However, when those networks are applied to different and unprecedented scenarios not included in the training set, they can suffer a drop in performance. For example, with robotic platforms in ever-changing and realistic environments, where always new sensory evidence is acquired, the performance of those models degrades. One solution is to make robots learn from their first-hand sensory data with self-supervision. This allows coping with the inherent variability of the data gathered in realistic and interactive contexts. To this aim, we propose a cognitive architecture integrating low-level perceptual processes with a spatial working memory mechanism. The architecture autonomously organizes the robot's sensory experience into a structured dataset suitable for human recognition. Our results demonstrate the effectiveness of our architecture and show that it is a promising solution in the quest of making robots more autonomous in their learning process. 

\end{abstract}

\section{INTRODUCTION}

The purpose of today's robotics is to support humans in diverse contexts which range from functional collaboration, as in manufacturing, to personal relationships, as in care-giving, education, and rehabilitation scenarios. Especially in contexts like the latter, human-robot interaction (HRI) needs to reach the next level of complexity, with robots capable not only to efficiently interact but also to account for the diversity among individuals \cite{tapus2008user}. To this end, it is necessary to personalize interactions based on the memory of shared experiences and let the robot adapt its behavior considering the history of previous interactions with humans \cite{belgiovine2020sensing, vignolo2020using}. Indeed, personalization and adaptation are keys to successful long-term HRI \cite{ahmad2017systematic}. These skills can improve users' engagement, making it last even after the novelty effect wears off \cite{szafir2012pay}. They can also help to establish a rapport and trust between the robot and users, for example by recalling shared memories  \cite{saez2015has} or by detecting changes in the partner's affective state \cite{Gordon2016}. \\

It becomes therefore increasingly necessary for robots to be able to reliably recognize the partners they have interacted with. 
Advances in artificial intelligence (AI), and more specifically machine learning (ML), provide today potentially powerful tools to perform human recognition.

Recently, several recognition algorithms using visual information and based on deep learning networks (e.g. face recognition algorithms) have been proposed achieving state-of-the-art results and establishing themselves as the preferred techniques to address such task \cite{Guo2019}. However, non-visual cues such as voice represent also important information for person recognition, especially when the person is not in the visual field of view or when vision is limited. Deep learning techniques have proved to be a robust tool also to address speaker recognition, achieving state-of-the-art results \cite{Sztaho}.
Nonetheless, several issues arise when applying AI methodologies to visual and non-visual information gathered in HRI frameworks. The problems are mostly caused by the scarcity of available data in these settings. 
Also, the diverse nature of the data stream, along with the inherent variability related to the robotic platform or the interactive context, could complicate the use of established AI solutions.

Deep Learning algorithms work well as long as the data distribution they were trained on is similar to the distribution they will be used on. For robots evolving in real-world scenarios, there is often a mismatch between the training and the target data distribution, mainly due to the use of different sensors (e.g., microphones and cameras) and to the noise in the environment or generated by the robots \cite{Sunderhauf2018}.
Considering the example of face recognition, with the advent of the COVID-19 pandemic, classical face recognition approaches hardly work with the previous solutions, because during interactions the faces are covered by masks. To cope with this issue, it would be necessary to collect new ad hoc datasets and to re-train the model, e.g. by using domain adaptation techniques.
However, this represents a strong limitation, as the collection and creation of these datasets are time-consuming as they frequently require human annotation.\\

In order to overcome this limitation and to increase the amount of relevant data that a robot could leverage, it is crucial to exploit the robot's embodiment in the real world. The advantage lays in perceiving directly the world around it and continuously updating the sensorial data from the environment.
The open challenge for the robotics community is therefore to design frameworks aimed at autonomously and effectively managing the data captured by the robot during recurrent interactions with humans (i.e., "its experience") so that it can use this information to learn in an autonomous and self-supervised way. The robot's ability to organize its own sensorial experience and to retrieve it in future contexts is of utmost importance for personalized, long-term HRI scenarios, where the amount of information grows continuously \cite{sandini2018social}. This is especially advantageous for interactions involving multiple people, behaving dynamically. \\


To this purpose, we propose a solution based on a cognitive architecture that exploits the robot's low-level perceptual mechanisms (e.g., sound localization and face detection) and working memory that endows the robot with the ability to autonomously organize its experience while interacting with humans.
Cognitive robotics focuses, among many capabilities, on designing and building cognitive models for robots that have the ability to learn from experience and from others. In particular,
the ability to retrieve experiences from memory as the context requires, and flexibly reuse this knowledge to select appropriate actions in the pursuit of inner goals 
\cite{Vernon} is fundamental for cognitive robotics. 

In this process, memory plays a key role, as it helps the agent to encode, store, and retrieve knowledge. In particular, working memory refers to the capacity to maintain temporarily a limited amount of information, which can then be used to support various work-dependent skills, such as learning, reasoning, planning, and decision-making processes. In other words, working memory is not just a retention mechanism, but it is also a framework that preserves transient goal-relevant pieces of information in the service of complex cognitive processes \cite{baddeley2012working}. The interest in exploring the role of different types of memory for human-robot interaction, with either embodied or virtual agents, is progressively gaining the attention of the community \cite{persiani2018working,  martin2021declarative}. However, its implementation in the robotic field for autonomous interactions with humans is still limited to few interesting researches \cite{stachowicz2011episodic, rothfuss2018deep, kasap2010towards}. 

In this work, we propose to use working memory and perceptual system together, to let the robot interact autonomously with multiple partners and cluster multi-modal data, by collecting them into a structured dataset suitable for the tasks of speaker and face recognition.
After introducing our architecture, we present its implementation in a game-like interaction scenario involving the humanoid iCub and three participants simultaneously and test it in a user study on a total of 12 participants. Finally, we report the results of testing different models for face and speaker recognition on our dataset, to analyze the pros and cons when implementing different methodological solutions on our limited and noisy dataset. 



\section{COMPUTATIONAL ARCHITECTURE}\label{section:architecture_section}

\begin{figure}[!th]
    \includegraphics[scale=0.3]{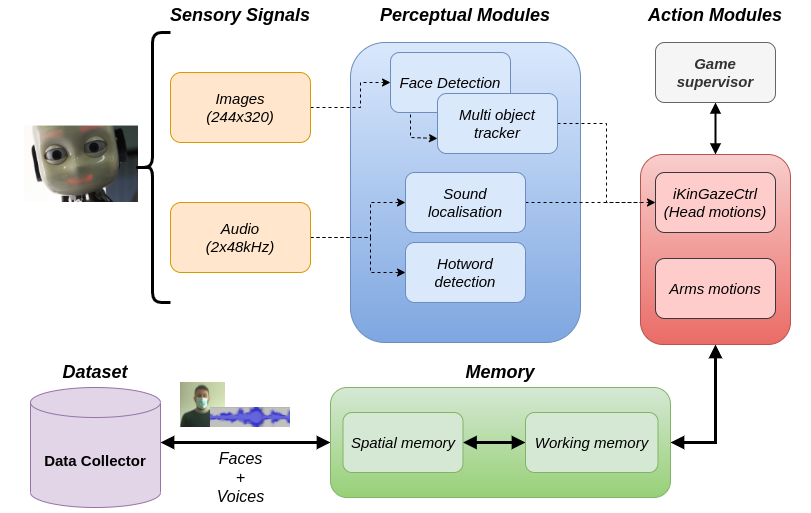}
    \caption[]{Representation of the computational architecture: The raw data from the iCub sensors are passed towards perceptual mechanisms which allow the extraction of information such as faces and voice location. The proprioceptive information of the robot along with the spatial memory system allow saving this information into a structured dataset.}
    \label{fig:architecture}
\end{figure}

We present in this section the different components of the architecture \textbf{(Fig. \ref{fig:architecture})} used to learn in a self-supervised way to recognize persons during a natural human-robot interaction.  All the different components are stand-alone modules and are integrated with the YARP middleware \cite{Metta2006} which provides a modular design of the cognitive architecture. \\

\subsection{Perceptual Modules}

\subsubsection{Sound Localization System (SLS)} 
\begin{figure}[!b]
    \vspace{-0.4cm}

    \centering
    \includegraphics[scale=0.85]{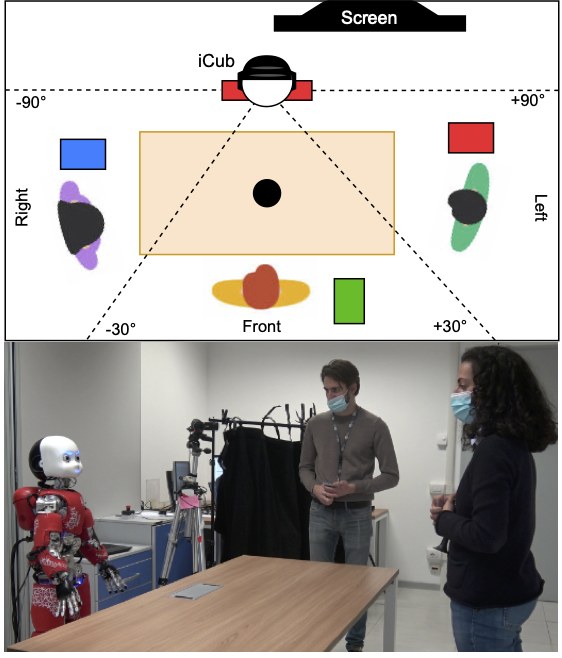}
    \caption[]{Top panel: Scheme of the experimental setup during the game phase. Bottom panel: Participants and the iCub robot while playing the history taboo game. The third participant was present to the right of iCub but is not visible in this image.}
    \label{fig:taboo_game}
\end{figure}

We used a sound localization system (SLS) to discern the incoming direction of auditory signals and thus clustering voices coming from sources differently placed in the space, with the assumption that the scene changes only in specific time instances. Once the relative position of the people is known in the space, the robot could infer the identity of the speaker by mapping the direction from which the voice comes to the information stored in the spatial memory. In this specific case, the direction of the sound has been discretized in three regions of the 180 degrees range of motion relative to the robot: left, right, and in front of it (see \textbf{Fig. \ref{fig:taboo_game}, top panel}).\\

\subsubsection{Hot-word detector}
To make the participant interact naturally and for a prolonged period without human intervention in the different phases of the game, we used Snowboy, an artificial intelligence software toolkit developed by KITT.AI for hot word detection\footnote[1]{https://github.com/Kitt-AI/snowboy}. We used the software to detect the words \textit{yes} and \textit{no} to automatically ask the players if the answer provided by another player was right or not. This allowed us to fully automatize the interaction without the need to rely on experimenter intervention or a Wizard-of-Oz approach. \\

\subsubsection{Face Detector}
To identify the participants interacting with the robot we use a deep learning-based face detection system developed for the iCub robot \cite{Gonzalez-billandon2020}. The face detector was trained in a self-supervised way and it is inspired by how infants learn to detect faces using audiovisual attentional mechanisms to guide their learning. \\

\subsubsection{Multiple Object Tracker (MOT)}
The face detection module allows for the localization of faces from the robot camera but cannot perform face identification. That is, from a sequence of frames, it cannot associate a currently detected face with a previously seen one. 
However, for the robot it is strictly necessary to keep track of an object's position in time and to distinguish it from other objects. 
To perform data association, we used a multi-object tracker algorithm based on the combination of a Kalman Filter and Hungarian algorithm. The combination of the modules is used for the tracking components and allows an accuracy comparable to state-of-the-art online trackers, but with real-time performance, which is crucial for robotic application \cite{Bewley2017}. We used the MOT module to associate for each participant interacting with the robot a unique tracker $T$, which is maintained during the whole task (see \textbf{Fig \ref{fig:mem_vers2}}).

\subsection{Action Module}
To perform the tracking of the faces and orient the head of the robot toward the localized sound source given by the SLS module output, we used the iKinGazeCtrl module: a controller for the iCub gaze capable of steering the neck and the eyes independently \cite{roncone2016cartesian}. To execute the different arms motion of the robot we used pre-programmed sequences of full-body movements in joint space to have perfectly replicable movement comparable across all the participants.  
The Game supervisor module managed the transition between specific states of the state machine, by triggering different robot behaviors (see \textbf{Fig. \ref{fig:state_machine}}).

\subsection{Spatial/Working Memory Module}
While interacting with others, the robot has the advantage to be embodied in the world and thus it can use alternative supervisory mechanisms to organize its incoming signals. 
For example, when a robot faces different speakers it can use the proprioceptive information of its head orientation to cluster the face and voice of the person. We used a spatial memory module representing working memory to allow iCub to map a limited number of persons interacting with it to their spatial positions according to an egocentric frame of reference.
We implemented the spatial working memory using a Dictionary data structure:
\begin{equation}
   D: Keys \rightarrow Values \cup \epsilon
\end{equation}

Where the \textit{keys} are the azimuth values of the robot head, the \textit{values} is a list of instances of trackers $T$ and $\epsilon$ is the empty set. In our implementation, we discretized the azimuth range [-90, 90] into three bins: \{left; center; right\} each 60 degrees wide (see \textbf{Fig \ref{fig:mem_vers2}}).

\subsection{Data collector Module}
The combination of the perceptual module with the working memory allows the robot to store and update the position of the persons in the space throughout the interaction. This information is used
to autonomously organize the sensory data (i.e., faces and voices) into a structured dataset: \\ 

   $Dataset: \{P_i: [(I_i, x_1,y_1, x_2, y_2);...;(I_n, x_1,y_1, x_2, y_2)] \\
   \cup [audio_i;...;audio_n]]\}$ \\

For each participant $P_i$ the module stores automatically the images and the associated bounding box of the face $(I_i, x_1,y_1, x_2, y_2)$ retrieved through the face detector, and correctly gives a label using the MOT and the memory module.

Moreover, when the robot hears a person speaking, the system stores the detected audio associating it to the corresponding person $P_i$, thanks to the combination of the SLS module, which identifies the sound direction, and the memory module that maps the proprioceptive information of the robot to the identity of the memorized players.

\section{MATERIALS AND METHODS}

\begin{figure*}[!th]

    \includegraphics[width=\textwidth]{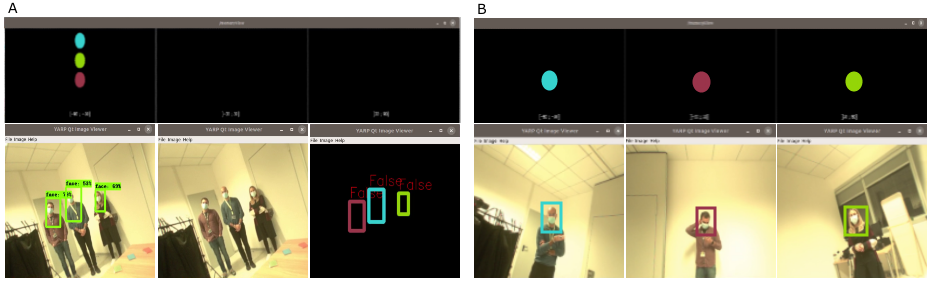}
    \caption[]{Demonstration of the Spatial Memory (top) and the MOT tracking module (bottom): (A) At the beginning of the interaction the three participants are memorized on the left position by the memory module. (B) After the positioning phase, the memory module is updated by the MOT module, and the players are successfully represented in their corresponding spatial positions according to the robot frame of reference.}
    \label{fig:mem_vers2}
\end{figure*}

\subsection{Participants}

The experiment was conducted with the support of 12 healthy participants that voluntarily agreed to participate to the data collection (8 males, 4 females, 27.25$\pm$1.48 years of age), divided in 4 groups of 3 participants each (1 female and 2 males).

\subsection{Experimental Design}

Participants were asked to play a modified version of the taboo game (where the words to be guessed, differently from the traditional taboo game, were historical figures) with iCub, a humanoid robot with a child appearance \cite{Metta2010}.

The objective of the game was to guess the identity of historical figures after one of the participants described them without pronouncing some keywords important for the identification (the "taboo words"). Also, participants scored proportionally to the number of cards correctly guessed.

The robot's role was to guide participants through the different stages of the game, without actively competing by describing cards or guessing (see \textbf{Fig \ref{fig:state_machine}}). It was explained to the participants that since iCub is a child robot, it could not  explain the characters in human history; however, it could learn by participating as a referee and supervisor of the game.

The choice of this experimental design stems from the need to design an engaging interaction that, at the same time, could well represent a real-world scenario where a robot supervises and modulates interaction in a group of multiple users, such as robot teachers in a class of students or robot therapists in a rehabilitation session.

\subsection{Protocol and State-Machine}

\begin{figure}[!th]
    \includegraphics[scale=0.47]{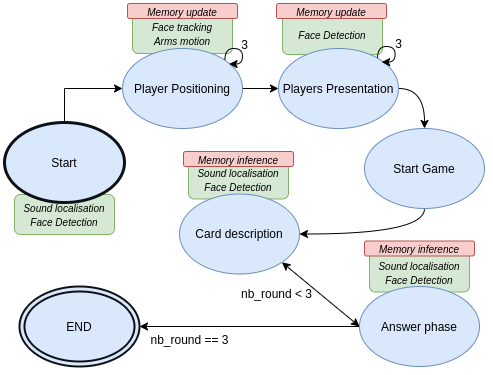}
    \caption[]{State machine followed by the robot during the game.}
    \label{fig:state_machine}
\end{figure}

Upon their arrival, participants were first brought to the office setting, where they were assigned a color-player label among blue, green, and red, that was later necessary during the game phase.  
They were also instructed that once in the experiment room, they should call iCub attention with their voice to make the interaction start. 
Indeed, using the \textit{SLS} module, iCub turned its gaze toward the direction of participants' voice and once it detected faces through the \textit{Face Detector} module (\textbf{Fig. \ref{fig:mem_vers2}A, bottom panel}) it started the interaction by welcoming them and updating the \textit{Spatial Memory} module with the position of the detected faces (\textbf{Fig. \ref{fig:mem_vers2}A, top panel}).

In the next phase \textit{Player Positioning} (\textbf{Fig. \ref{fig:state_machine}}), iCub invited one color-player at a time (in random order) to approach the table placed in front of it and to take the deck of cards of the corresponding color, containing the historical figures with the associated taboo words. Then, it asked the participant to take other positions in the room by pointing with its arm on its right, front, or left. 
In this phase, iCub associated the color-player with the tracked face, which was stored and labeled to be organized in the database through the \textit{Data Collector} module (\textbf{Fig. \ref{fig:mem_vers2}A, bottom panel}). As the participants moved towards the position, iCub followed them with its gaze by using the \textit{MOT} module. When the robot detected participants stopping, it updated their position in the space through the \textit{Spatial/Working Memory module}. In this way, each color-player was mapped to a position relative to the robot reference frame (\textbf{Fig. \ref{fig:mem_vers2}B, top panel}).

Then, the \textit{Players Presentation} phase began and iCub addressed one player at a time to ask them their name and to invite them to present themselves in a pre-set time of 20 seconds.
This phase strongly relied on the role of memory and had multiple purposes: i) to automatically retrieve the person's name for the creation of the database and the recognition phase, replacing the player-color label with it, and ii) to prepare a framework already suitable for future one-shot learning scenarios, where one needs to have initial reference samples to implement on-the-fly person-recognition models.

At this point, the actual history taboo game started (\textit{Start Game} phase). On a screen placed behind the robot, a colored circle appeared giving the signal to the player with the corresponding color to prepare to describe the card. After few seconds, a 30 seconds timer started, during which the player had to describe the card to the other players. 

During this \textit{Card Description} phase, we intentionally deprived the iCub of a priori knowledge of the players' turns. This replicates common interactive contexts in real-world scenarios where the robot has to orient its attention in the scene relying exclusively on its perceptual mechanisms. In the game, it turned its gaze toward the direction inferred by the \textit{SLS} module, it associated the detected voice and face to the color-player label by querying its spatial memory, and it stored them in memory with the corresponding label using the data collector module. At the end of the timer, iCub interrupted the player by saying: "Time's up. I wonder if anyone has figured out what this is all about?".

In the \textit{Answer phase}, the player who wanted to answer, at the iCub signal had to grab the buzzer placed in the middle of the table and call iCub attention with his/her voice.
Once iCub turned its attention to one of the players (with the same mechanisms described in the previous phase), it invited the person to give the answer, and then it turned again to the previous player to verify whether it was correct or not. 
To make the interaction completely autonomous, we exploited the \textit{Hot-word detector} module to detect the words \textit{"yes"} and \textit{"no"}, as explained in section \ref{section:architecture_section}. Also, during the interaction, iCub showed different facial expressions depending on the context through LEDs representing its eyebrows and mouth. 
The game was repeated until each player had described a total of three cards. 

Each session of interaction lasted about 30 minutes. The state-machine flow and robot behaviors were completely autonomous and triggered by events that relied solely on the robot's perception (e.g., sound or face detection) and game dynamics (e.g., timers). The experimenter monitored the correct progress of the experiment from a different room, ready to intervene only in case of problems.

\section{DATA ANALYSIS}

\subsection{Database generation}

During each interaction, the architecture stored autonomously the different data points of faces and voices by associating them with the corresponding player by using the color labels coding. In a post-processing phase, the \textit{Data Collector} module organized the visual and auditory data by creating a structured dataset as described in previous sections. The dataset is suited to perform speaker and face recognition through supervised-learning. In fact, the dataset was populated by creating labels corresponding to the name of each player, extracted from the audio files saved in the \textit{Players Presentation} phase, by using the Google-Speech recognition API. For each of these labels (12 in total), two different structured datasets were created: $Dataset\_faces$ and $Dataset\_voices$, described in detail below.  In other words, the \textit{Data Collector} module reorganizes the robot's temporary experience into a structured knowledge about people who interacted with it, creating a memory made up of unique names, faces, and voices.\\ 

Through the experiment, we collected a total of 1320 seconds of audio files (110 seconds on average for each participant), and 1200 samples of face images (93 on average for each participant, see Table \ref{table:dataset_details}). 
The audio recorded presented an SNR equal to 9 Db due to the high ego-noise of the robot produced by the fans in its head (q realistic and undesired characteristic of typical robotic applications). 
We pre-processed each face according to the face alignment procedure for the face recognition network \cite{Schroff}. The face alignment procedure consists of centering the faces bounding boxes and resizing them to a fixed dimension of 180x180 pixels (see \textbf{Fig  \ref{fig:example_face_dataset}}). 


\subsection{Triplet-Loss network}
\label{section:triplet_loss_model}
    When addressing the task of face or speaker recognition one solution is to rely on triplet-loss trained deep networks. The triplet loss approach comes from the name of the loss function used to train a deep network to produce euclidean embeddings with the goal to minimize the distance between an anchor and a positive sample and maximize the distance between the anchor and a negative sample of a different identity \cite{Weinberger2009}.
    Moreover, as the triplet loss enforces negative samples to be distant in the embedding space, the network can be used to identify previously unseen persons by using similarity or distance measures.
    We investigated the performance of pre-trained triplet-loss networks on our collected datasets. 
    For speaker recognition, we used a Resnet network trained on the LibriSpeech dataset \cite{panayotov2015librispeech}.  We used the network to create a database of embeddings ($D_{emb-voices}$) for each person in the  $Dataset\_voices$.  
    For the faces, we used the Facenet network \cite{Schroff}, trained on the large-scale face dataset VGGFace2 \cite{Cao2018}, to produce the database of face embeddings ($D_{emb-faces}$) from the  $Dataset\_faces$.
    
    To perform the identification, we used a KNN algorithm with $k=1$ to quickly find the nearest elements from the embeddings database. 
    We then used the returned distances and a threshold $t$ (empirically chosen) to classify the person as the nearest embedding class (distance $> t$) or as an unknown person (distance $< t$).

\subsection{Multi classification}
\label{section:multiclassification_models}

We also investigated the performance of a multi-classification approach on our datasets. For the speaker recognition, we trained a deep network based on stacked convolutional layers with gammatonegrams features as input. Gammatonegrams decompose a signal by passing it through a bank of gammatone filters equally spaced on the ERB scale and were designed to model the human auditory system. We opted for gammatonegrams as they have proven to be robust to noisy environments \cite{maganti2011auditory} and thus suited to deal with robot data. We pre-processed the $Dataset\_voices$ by taking chunks of audio of 1 second with a hop-length of 250ms and we discarded chunks having less than 80\% of voice data using the Google voice activity detector. Finally, we generated the gammatonegrams on the chunks using  a bank of  128 filters and stacked horizontally the two channels, giving a final feature size of 128x192. Instead, for the face recognition model  we opted for  a transfer learning approach and fine-tuned the Facenet network to perform multi-classification task.

\subsection{Test Dataset}
\label{section:test_dataset}

\begin{figure}[!th]
    \centering
    \includegraphics[scale=0.35]{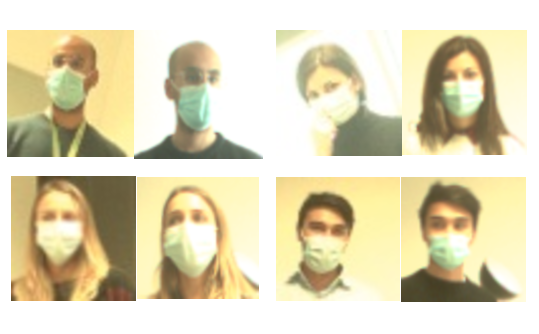}
    \caption[]{Samples from the collected dataset: For each person, the right photo is a sample from the training dataset (collected during the game interaction) and the left one from the test dataset.}
    \label{fig:example_face_dataset}
\end{figure}

To test the proposed models for voice and face recognition, we asked participants to come back after few days in order to acquire additional data via the robot's sensors. At this stage, there was no actual interaction between the participants and the robot, as our goal was to demonstrate the robustness of our approaches using a dataset acquired under different conditions of the environment (lighting, etc..) and of participants (see as example \textbf{Fig. \ref{fig:example_face_dataset}}). The approach better approximates the application contexts in which the model will be deployed. 
 We also collected face and voice data of 4 people who did not participate in the experiment, to test the accuracy of the triplet-loss models in recognizing unseen samples. For both the faces and the voices, we applied the same pre-processing described in sections \ref{section:multiclassification_models}, \ref{section:triplet_loss_model}. The final test-set resulted in 6 audio samples (of 1 second) and 10 faces per person (see \textbf{Table \ref{table:dataset_details}}).

\section{RESULTS}

\vspace{-0.4cm}

\subsection{Architecture and data collection}

\begin{table}[!ht]

\caption{Details of the collected data during the game interactions using our proposed architecture}
\centering
\label{table:dataset_details}
\begin{tabular}{l  c c}
\toprule
\textbf{Dataset type} & \multicolumn{1}{c}{\textit{Avg data points per persons}} & \multicolumn{1}{c}{\textit{Avg size}}\\ 
\midrule
Faces train set & 93  & 32x57  \\
Faces test set & 10  & 34x58 \\

 \midrule

Voices train set & 110s   & - \\
Voices test set & 6s  & - \\
\bottomrule 
\end{tabular}
\end{table}

Our architecture proved to be efficient in autonomously labeling and clustering visual and auditory data during a dynamic interaction with multiple people. On a total of 36 trials, the SLS module showed an accuracy of 97$\%$. The robot failed in predicting sound direction - and consequently the player label associated to that direction by the memory module - only once across all trials. Regarding the automatic creation of the datasets, using the Google speech recognition API we successfully extracted the different names of the participants gathered during the \textit{Player Presentation} phase with an accuracy of 83.3\%. Indeed, out of the twelve participants who played the game, one player was not correctly recognized (\textit{Parco} instead of \textit{Marco}) and one player could not be correctly processed by the software and was therefore automatically labeled as "unknown".

\begin{figure}[!th]
    \vspace{-0.3cm}

    \centering
    \includegraphics[scale=0.2]{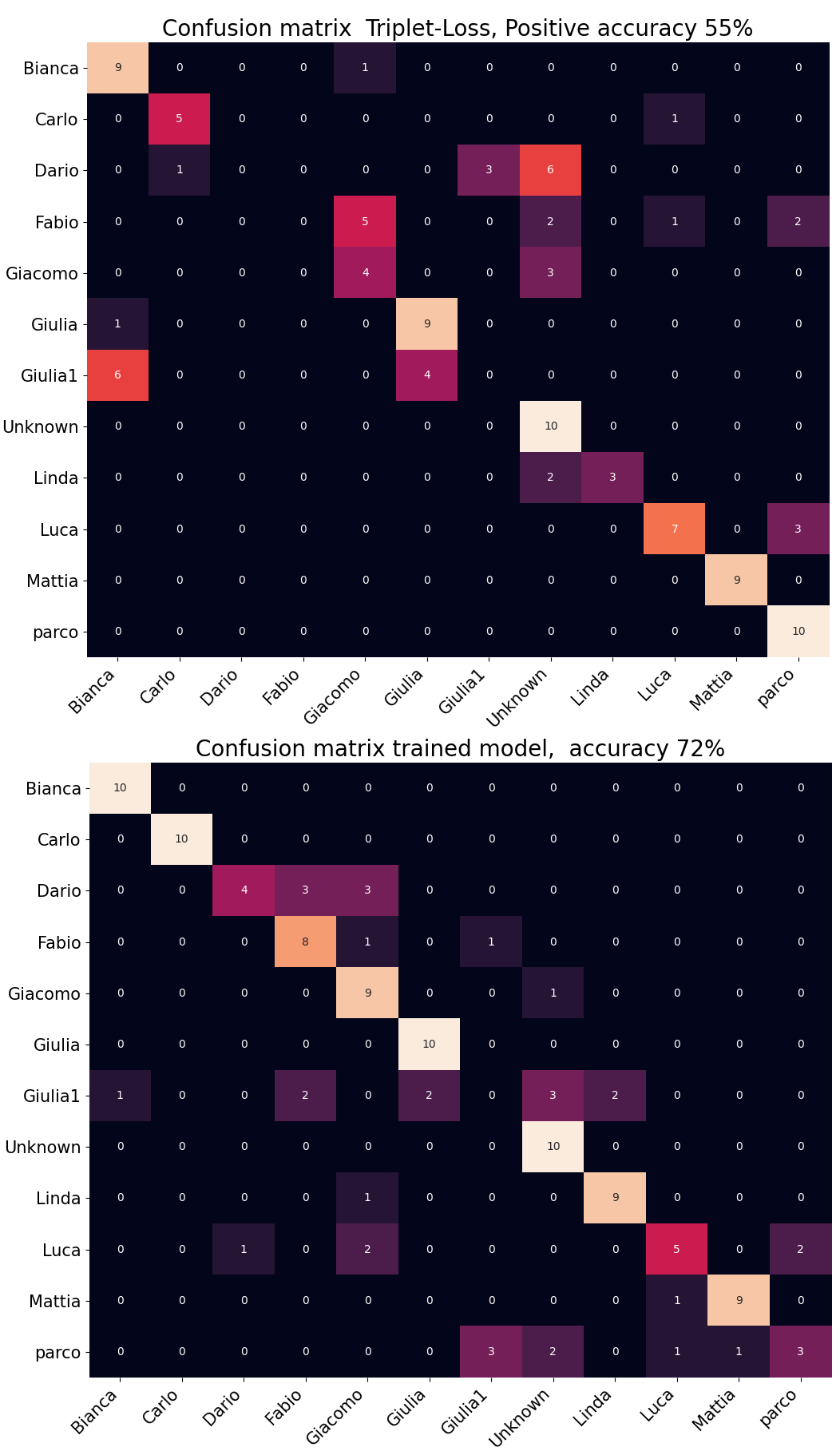}
    \caption[]{Face Recognition classification results. Top: Triple-loss model positive accuracy. Bottom: Re-trained model with the data collected during the interaction using transfer learning.}
    \label{fig:face_confusion_matrix}
\end{figure}

\subsection{Triplet-loss networks}
To asses the performances of the triplet-loss networks for both the faces and voices, we computed the positive accuracy and negative accuracy:

\vspace{-0.5cm}

\begin{align*}
    Positive \;  accuracy = TP / (TP + FN)\\
    Negative \; accuracy = TN / (TN + FP) 
\end{align*}

We tested both triplet-loss networks on the gathered test-set to compute the positive accuracy and further tested on unseen participants to compute the negative accuracy (see \ref{section:test_dataset}). We used the Euclidian distance and fine-tuned the threshold $t$ to balance between the positive accuracy and negative accuracy. For the face recognition, using a $t=0.4$ the triplet-loss network achieved a positive accuracy of 55\% (see \textbf{Fig \ref{fig:face_confusion_matrix}}) and a negative accuracy of 73\%. Instead, for the voices the best compromise found between the positive accuracy and the negative accuracy was 30.3\% and 37.5\% respectively using $t=1.7$ .

\begin{figure}[!th]
    \includegraphics[scale=0.2]{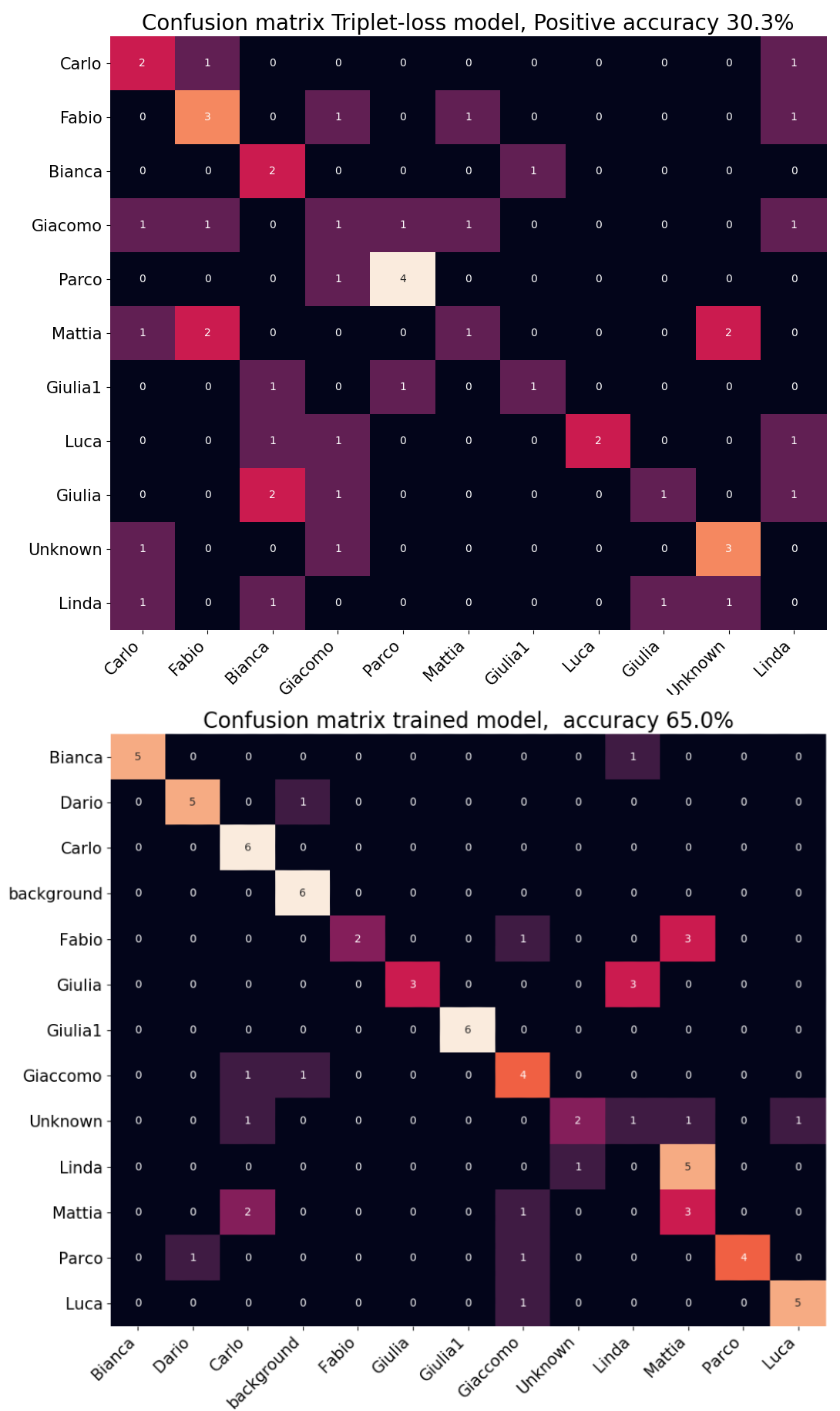}
    \caption[]{Speaker recognition classification results. Top: Triplet-loss model test positive accuracy. Bottom: New model based on gammatonegrams test accuracy}
    \label{fig:voice_confusion_matrix}
\end{figure}

\subsection{Multi-classification networks}

 We re-trained the Facenet network for a multi-classification task on the 12 persons present in our dataset. The re-trained network achieved on the test set (see  \ref{section:test_dataset}) an accuracy of 72\% (\textbf{Fig \ref{fig:face_confusion_matrix} bottom}) against a  chance level of 8\%. For the audio, we re-trained the model to classify the 12 persons plus an extra class, background, which we obtained by recording the ego-noise of the robot. The final model reached a total accuracy of 65\% on the voice test set.

\section{DISCUSSIONS}

In this work, we proposed an architecture to enable robots to organize autonomously multimodal data gathered while interacting with a group of human partners. The objective was to efficiently synthesize the knowledge necessary to replicate a person recognition system and build natural HRI interactions based on the history of users' experiences. 
To achieve this, we built an architecture based on basic perceptual mechanisms: the localization of sounds in space, the detection of faces and  the tracking of moving objects. These are primary cognitive processes that can be exploited by any embodied agent in real-life contexts. These basic perceptual mechanisms were integrated with a working spatial memory system. The spatial memory allowed the robot to associate each user to a specific spatial location relative to the robot's egocentric frame of reference and consequently to infer the user's identity from the sensorial data coming from the specific environmental region.

We demonstrated that the proposed architecture worked efficiently in a prolonged game-like interaction between the robot and multiple persons. 
Using our architecture, we successfully created a dataset of people who interacted with the robot without the need for human intervention. The robot was not able to correctly recognize the name of participants in few situations, which can be solved in the future by enriching the interaction with advanced skills (e.g., by letting the robot ask for the user's name again when the output of the software is not reliable).

We used the obtained dataset to develop a face and speaker recognition system exploring two different approaches: pre-trained triplet-loss models and re-trained multi-classification models. The processing of data and training of the models relied on post hoc analysis for a preliminary evaluation of the recognition methods, but in the future, they could be implemented in a fully automated pipeline process using DevOps tools. 

The obtained results demonstrated a higher accuracy for the re-trained models, which was to be expected in light of the data collected by the robot. Indeed, the low-resolution faces images, the presence of the mask, and the low SNR of the audio explained the poor performance of the triplet-loss networks which were trained on very different sets.  For example, the LibSpeech contains clean speech of people without noise and in the VGGDataset faces with masks were not present. 

This demonstrates the advantage of using the data directly collected by the robot during its interaction with the world and of developing frameworks able to manage and manipulate the on-line robot experience. Thanks to this, the robot can learn through a self-supervised strategy while interacting with humans and with the world. Indeed, the effectiveness of such approaches is likely to improve as the number of interactions, and thus of the available data, increases. Moreover, with more data available it could be possible to train a triplet-loss network from scratch and fully take advantage of this approach.  

Considering the limitations of our dataset collected during a real interaction, as we expected, the obtained results are hardly comparable to the state-of-the-art face/speech recognition models. Nonetheless, this was not our primary goal. Rather our intent was to demonstrate the robustness of our approach to the variability of natural interactions with humans. \\

Such variability may derive from several, often unpredictable sources such as the different conditions of the environment, the inherent noisy nature of data, or the dynamic changes of the interactive context. Consider for example the limitations of trying to implement a facial recognition system during the pandemic, where the faces covered by masks are difficult to distinguish. 
In these scenarios, it is important that the robot exploits its direct experience to learn how to cope with these situations and adapt accordingly, for example by relying more on the voice data.

To conclude, our proposed architecture represents a novel contribution to the development of robots capable of autonomously and efficiently managing their sensory experience to build a multimodal memory of the people they have interacted with. This memory could be improved so that a robot could autonomously learn to manage the amount of data collected, by synthesizing it and by selectively removing the non-necessary information, similarly to what humans do. The proposed framework proved to be accurate and functional to manage a fully autonomous interaction between the robot and a group of people. 
Moreover, our approach facilitates building personalized and adaptive HRI. Indeed, having a history of the facial, auditory, and contextual features of users in a long-term memory could allow the robot to relate the current experience to past experiences and for example to monitor changes in the partner's inner and affective state, health, or mood - adapting its behavior accordingly.






\section*{ACKNOWLEDGMENT}

A.S. is supported by a Starting Grant from the European Research Council (ERC) under the European Union’s Horizon 2020 research and innovation programme. G.A. No 804388, wHiSPER.


\bibliographystyle{IEEEtran}
\bibliography{IEEEabrv,ref.bib}

\end{document}